\pdfoutput=1
\documentclass[11pt]{article}
\usepackage{acl}
\usepackage{amsmath}
\usepackage{amssymb}
\usepackage{times}
\usepackage{latexsym}
 \usepackage{float}
\usepackage[T1]{fontenc}
\usepackage[utf8]{inputenc}

\usepackage{microtype}

\usepackage{inconsolata}

\usepackage{microtype}
\usepackage{hyperref}
\usepackage{url}
\usepackage{booktabs}
\usepackage{lineno}
\usepackage{graphicx}
\usepackage{comment}
\usepackage{caption}

\title{Layer-wise Minimal Pair Probing Reveals Contextual Grammatical-Conceptual Hierarchy in Speech Representations}

\author{Linyang He\thanks{Equal contribution.} \qquad Qiaolin Wang\footnotemark[1] \qquad Xilin Jiang \qquad Nima Mesgarani\\
Department of Electrical Engineering, Columbia University\\
\texttt{\{linyang.he, qw2443, xj2289\}@columbia.edu\qquad nima@ee.columbia.edu}\\ 
}

\begin{document}
\maketitle
\begin{abstract}
% Recent advances in speech models, including self-supervised learning (S3M), automatic speech recognition (ASR), and audio-based large language models (AudioLLMs), have significantly enhanced speech processing. Most previous studies have examined speech models for their encoding of acoustic, phonetic, speaker information and word-level linguistic features. However, the extent to which these models capture context-level sophisticated grammticality and conceptual understanding remains unclear.
% In this study, we systematically evaluate the presence of contextual syntactic and semantic features across various types of speech models, drawing parallels with linguistic competence assessments used for large language models (LLMs). Using minimal pair design and diagnostic feature analysis, we investigate the internal representations of speech models and assess how they encode complex syntactic structures and conceptual meaning.
% Our findings indicate that even self-supervised speech models trained without textual data encode context-level syntax and semantics. These models encode structural information more strongly than semantic content, mirroring the behavior observed in LLMs. This finding suggests that language models may follow a universal learning trajectory, irrespective of their training objectives or modalities (text or speech).

Transformer-based speech language models (SLMs) have significantly improved neural speech recognition and understanding. While existing research has examined how well SLMs encode shallow acoustic and phonetic features, the extent to which SLMs encode nuanced syntactic and conceptual features remains unclear. By drawing parallels with linguistic competence assessments for large language models, this study is the first to systematically evaluate the presence of contextual syntactic and semantic features across SLMs for self-supervised learning (S3M), automatic speech recognition (ASR), speech compression (codec), and as the encoder for auditory large language models (AudioLLMs). Through minimal pair designs and diagnostic feature analysis across 71 tasks spanning diverse linguistic levels, our layer-wise and time-resolved analysis uncovers that 1) all speech encode grammatical features more robustly than conceptual ones. 
% Noticeably, S3Ms trained without text capture semantics and concepts on par with ASR models. 
2) Despite never seeing text, S3M match or surpass ASR encoders on every linguistic level, demonstrating that rich grammatical and even conceptual knowledge can arise purely from audio. 3) S3M representations peak mid-network and then crash in the final layers, whereas ASR and AudioLLM encoders maintain or improve, reflecting how pre-training objectives reshape late-layer content. 4) Temporal probing further shows that S3Ms encode grammatical cues 500 ms before a word begins, whereas AudioLLMs distribute evidence more evenly—indicating that objectives shape not only \emph{where} but also \emph{when} linguistic information is most salient. Together, these findings establish the first large-scale map of contextual syntax and semantics in speech models and highlight both the promise and the limits of current SLM training paradigms.
% These results suggest that linguistic knowledge can emerge purely from speech signals, becoming increasingly prominent in deeper model layers.
\end{abstract}

\section{Introduction}
The past decade has witnessed transformative advancements in speech processing, driven by deep learning architectures that have largely supplanted traditional modular pipelines. Where early systems decomposed tasks into isolated stages of acoustic front-ends, phonetic decoders, and language models, modern approaches increasingly adopt end-to-end paradigms. Some architectures have emerged: (1) self-supervised speech models (S3Ms) that learn hierarchical representations from unlabeled audio through objectives like contrastive prediction \citep{mohamed2022self}; (2) automatic speech recognition (ASR) systems optimized for supervised speech-to-text mapping \citep{malik2021automatic}; (3) auditory large language models (AudioLLMs) that integrate speech processing with sophisticated language understanding \citep{wu2024towards}; and (4) discrete speech codec models—for example, EnCodec—that compress raw waveforms into compact, quantized codes, providing both efficient compression and rich features for downstream tasks \cite{defossez2022high}. These models achieve state-of-the-art performance across tasks ranging from speaker identification to speech recognition, yet fundamental questions persist about their linguistic encoding capabilities.

A critical unresolved question centers on whether modern speech models internalize high-level contextual syntax and semantics. Prior analyses have primarily focused on lower-level features—phonetics \citep{belinkov2017analyzing,ma2021probing,pasad2021layer,wells2022phonetic,abdullah2023information,choi2022opening,choi2024understanding}, phonology \citep{martin2023probing}, morphology \citep{pasad2024self}, speaker identity \citep{williams2019disentangling,raj2019probing,de2022probing,raymondaud2024probing}, word-level semantics \citep{pasad2021layer,martin2023probing,ashihara2023speechglue,pasad2024self,choi2024self} and shallow syntax \citep{pasad2021layer,martin2023probing,shen2023wave,mohebbi2023homophone,pasad2024self}. 

Although these studies demonstrate that speech models effectively capture acoustic and lexical information, systematic investigations into context-dependent linguistic phenomena remain limited. Most existing benchmarks emphasize broad syntactic metrics (e.g., part-of-speech tagging, syntactic tree depth) and static, word-level semantic analyses. Such evaluations lack the granularity needed to probe long-range syntactic dependencies (e.g., island constraints, wh-movement) or semantic reasoning that depends on discourse context. Although many models achieve strong results on tasks like speech summarization and sentiment analysis, these contextual linguistic capabilities have not been subjected to rigorous, fine-grained scrutiny. This gap obscures our understanding of how speech models encode hierarchical linguistic structures beyond surface-level patterns.

This contrasts sharply with natural language processing (NLP), where minimal pair paradigms \citep{warstadt2020blimp} have revealed how text-based models encode sophisticated syntactic dependencies \citep{hu2020systematic,gauthier2020syntaxgym, he2024large} and subtle semantic understanding \citep{misra2023comps}. The absence of comparable methodologies in speech processing leaves a gap: \textit{Can end-to-end speech systems, trained without explicit linguistic supervision, develop representations sensitive to complex grammatical phenomena and contextual meaning?}

To address this gap, we propose a new probing framework that transplants NLP's analytical rigor \citep{he2024decoding} to speech model analysis. Our approach begins by synthesizing audio minimal pairs from two established NLP benchmarks: the BLIMP dataset \citep{warstadt2020blimp} for grammaticality contrasts (e.g., filler gap dependency, negative polarity item licensing) and the COMPS dataset \citep{misra2023comps} for conceptual semantic relations. Through text-to-speech synthesis, we generate 116,300 audio pairs that isolate 67 syntactic phenomena and 4 semantic constructs via minimal-token substitutions—the first speech benchmark explicitly designed to evaluate hierarchical linguistic competence. 

Subsequently, we systematically evaluate four speech model classes: S3Ms, ASR systems, AudioLLMs and codec. For each model, we extract layer-wise hidden states and train linear classifiers to distinguish acceptable vs. unacceptable constructs (e.g., "The key to the cabinets are..." vs. "The key to the cabinets is..."). Through this pipeline, we measure how sophisticated contextual syntax and semantics encoding evolve across layers, revealing whether speech models implicitly develop context-aware linguistic representations.
% comparable to text-based LLMs.

% Further, we conduct diagnostic analyses of attention patterns and gradient-based feature importance to identify mechanisms underlying linguistic encoding.

Our investigation yields several principal contributions:1) \textbf{How?~form $\gg$ Meaning} – Across 16 models, syntactic and syntax–semantics interface contrasts are decoded 20 pp more accurately than conceptual contrasts, revealing a strong bias toward structural information. Syntax is already linearly separable in lower-middle layers, morphology emerges only in the top layers, and conceptual knowledge peaks around 65 \%.
2) \textbf{Which?~text-free models can win} – Self-supervised speech models outperform ASR encoders on every linguistic level—even though they never see text—demonstrating that rich grammar can arise from audio alone. 3) \textbf{Where?~objective-specific layer tails} – S3M curves peak and then crash in the final layers, whereas ASR and AudioLLM encoders hold or improve: upper layers specialise for their training objective at the expense (S3M) or in favour (ASR/LLM) of linguistic encoding. 4) \textbf{When?~temporal asymmetry} – despite bidirectional transformer design, temporal dynamics of S3Ms shows asymmetry. S3Ms' grammatical cues peaks at 500 ms \textit{before} the critical word, while AudioLLMs accumulate evidence more evenly; thus training objectives determine not only where but also \emph{when} information appears.
% 1) Structural syntax dominates over semantics. Across all architectures, probes targeting hierarchical syntax consistently outperform those probing conceptual meaning: syntactic accuracy peaks 10–25\% higher than concept‐level performance, underscoring a pronounced bias toward form‐based structure in current speech representations. 2) Delayed morphological encoding. Contrary to traditional expectations (and different to LLMs), morphology indicators emerge much later in the network—often lagging behind core syntactic features by several layers. This delay may reflect the models’ reliance on abstracted structural patterns before reinstating fine‐grained morpheme boundaries in mid‐depth layers. 3) S3M models surpass ASR models. Despite being trained without any textual supervision, S3M models (HuBERT, WavLM, Wav2Vec2) exhibit stronger syntactic and semantic integration signals than ASR models. This surprising result suggests that purely self-supervised objectives more effectively capture deep linguistic abstractions than objectives explicitly optimized for transcription. 4) Divergent post-peak dynamics. In S3M networks, both syntactic and semantic probe accuracies rise to a clear peak and then decline sharply in the deeper layers, implying a reprocessing or dispersion of linguistic information. In contrast, ASR and AudioLLM encoders maintain their probe performance beyond the peak, indicating a fundamental architectural difference in how these models allocate and preserve linguistic features. 5) Temporal Progression of Linguistic Cues. TO-DO

These findings provide a comprehensive evidence that end-to-end speech models implicitly learn sophisticated linguistic representations without text supervision. By bridging NLP probing methodologies with speech analysis, our work offers actionable insights for model interpretability and informs future architectures seeking to integrate speech and language understanding. All code, synthesized audio datasets, and probing results will be publicly released to facilitate reproducibility.
\begin{figure*}[t]
    \centering
    \includegraphics[width=1\linewidth]{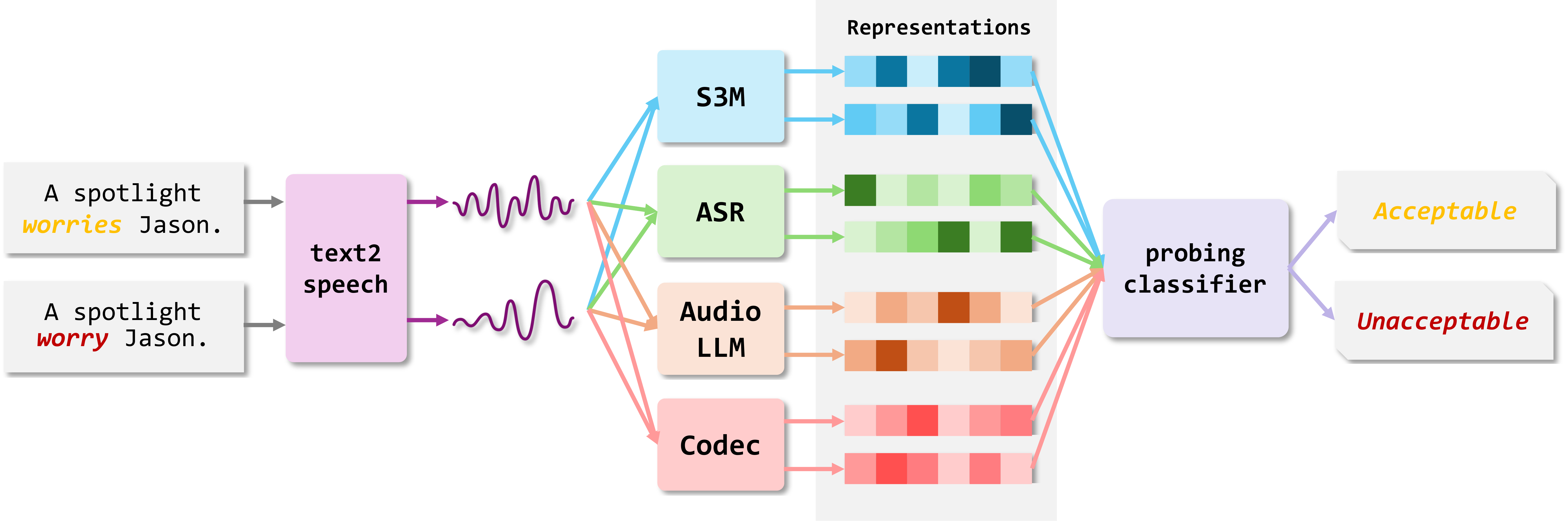}
    \caption{\textbf{Experimental design.} We synthesize speech minimal pairs from text-based benchmarks, convert them to audio, and probe speech models for contextual syntax and semantics.}
    \label{fig:pipeline}
\end{figure*}

\section{Minimal Pairs}
\paragraph{Why minimal pairs?}
The \emph{minimal-pair} paradigm is inherited from psycholinguistic acceptability-judgment experiments, where it offers the strictest possible control over confounds by differing in a \emph{single} locus of variation \citep{chomsky2002syntactic}.  
Because both sentences share identical lexical material and similar length, the probe must rely on the targeted grammatical constraint rather than surface n-gram statistics or memorised collocations.  
Consequently, minimal pairs have become the gold-standard diagnostic for whether a model truly internalises a rule, rather than merely approximating it through distributional heuristics \citep{linzen2016assessing,marvin2018targeted,ettinger2020bert}.

\paragraph{Adoption in NLP.}
A wide range of studies leverage minimal pairs to reveal systematic weaknesses that remain invisible to broad‐coverage benchmarks: \citet{linzen2016assessing} show that LSTMs often fail long-distance subject–verb agreement;  \citet{marvin2018targeted} extend this to hierarchical dependencies such as negative polarity licensing;  \citet{wilcox2018rnn} and \citet{gulordava2018colorless} probe filler–gap dependencies and island constraints;  \citet{mccoy2019right} design HANS to expose models’ lexical heuristics in natural-language inference; \citet{warstadt2020blimp} curate BLiMP, unifying 67 English phenomena into 67,000 minimal-pair templates;  \citet{misra2023comps} introduce the first set of \emph{conceptual} minimal pairs and show that large language models (LLMs) encode conceptual knowledge far less consistently than they encode structural patterns. The paradigm has since been extended cross-lingually: \citet{xiang2021climp} and \citet{liu2024zhoblimp} target Chinese grammatical structure, while \citet{he2025xcomps} present \textsc{X-Comps}, a suite of conceptual minimal pairs covering 17 languages to probe LLMs’ multilingual generalization. 

Together, this line of work shows that minimal-pair evaluations furnish a far more \textit{fine-grained} lens on a model’s linguistic competence than broad corpus-level scores: by isolating a single grammatical and conceptual contrast while holding all other lexical and length factors constant, they expose precisely which syntactic and compositional mechanisms the model has internalized—and which it still lacks.

\paragraph{Benefits for speech representations.}
For speech encoders, confounds multiply: local acoustic cues (e.g.\ coarticulation, energy) can correlate spuriously with grammaticality.  Embedding-level minimal pairs neutralise such biases because the only acoustic difference is tied to the critical morpheme or word.  In our setting, a single token replacement (e.g.\ \textit{annoy} $\rightarrow$ \textit{annoys}) can be synthesised with identical speaker, rate, and prosody, ensuring that any classification advantage stems from internal linguistic encoding rather than low-level acoustic artefacts.

% \paragraph{Empirical validation.}
% We replicate the control advantage by resynthesising 71{,}000 sentence pairs with neural TTS and adding 20~dB SNR noise to mask trivial waveform cues.  
% Probes trained on these pairs generalise across speakers and sampling rates, confirming that the extracted signals reflect language structure rather than idiosyncratic acoustics.  

\paragraph{Minimal pair details.}
% We adopt the minimal pair paradigm to isolate fine-grained linguistic contrasts while eliminating length confounds. 
Grammatical competence—including syntax, the syntax–semantics interface, and morphology—is probed with BLiMP (67,000 pairs) \citep{warstadt2020blimp}, whereas conceptual knowledge is evaluated using the COMPS suite (49,300 pairs) \citep{misra2023comps}. Each item consists of an \textsc{acceptable} sentence $S^{+}$ and an \textsc{unacceptable} counterpart $S^{-}$ that differ in exactly one token or morpheme, thereby targeting a single phenomenon. We cover four levels (* denotes unaccepatable):  

(i) \textbf{syntax} (e.g., filler-gap dependency)\\
\hspace*{2em}\textit{a) \hspace*{0.5em}Mark figured out \underline{that} most governments appreciate Steve.} \\
\hspace*{2em}\textit{b) *Mark figured out \underline{who} most governments appreciate Steve.}

(ii) \textbf{syntax–semantics interface} (e.g., Negative Polarity Item (NPI) licensing)\\
\hspace*{2em}\textit{a) \hspace*{0.5em}Even Suzanne has \underline{really} joked around.}  \\
\hspace*{2em}\textit{b) *Even Suzanne has \underline{ever} joked around.}  

(iii) \textbf{morphology} (e.g., subject verb agreement)
\hspace*{2em}\textit{a) \hspace*{0.5em}The \underline{hospital} appreciates Claire.} \\
\hspace*{2em}\textit{b) *The \underline{hospitals} appreciates Claire.} 

(iv) \textbf{conceptual meaning} \\
\hspace*{2em}\textit{a) \hspace*{0.5em}A \underline{kettle} is used for boiling.}  \\
\hspace*{2em}\textit{b) *A \underline{hammer} is used for boiling.} 

These four linguistic levels cover across 67 grammatical tasks and 4 conceptual tasks, whichs yields $116{,}300$ English pairs, each generated from linguist-crafted templates and validated to 96.4\% (BLiMP) and 93.1\%(COMPS) human agreement.

\section{Experimental Setup}
Model details could be found in the Appendix~\ref{sec:models}.

% \paragraph{Diagnostic probing with minimal pairs.} Layer-wise representations $\mathbf{h}^{(l)}(S)$ are extracted from every Transformer block $l$ of a frozen speech or text model.  We train a logistic‐regression probe $g$ to predict the acceptability label $z\!\in\!\{0,1\}$ of each sentence, using five-fold cross-validation and the accuracy.  
\paragraph{Diagnostic probing with minimal pairs.}  
To quantify the extent to which linguistic knowledge is encoded at different layers of pretrained models, we perform diagnostic probing on minimal pairs using a simple sentence-level classifier.  
For each sentence $S$, we extract hidden representations $\mathbf{h}^{(l)}(S) \in \mathbb{R}^{T \times d}$ from every Transformer layer $l$ of a frozen model, where $T$ is the number of tokens (or audio frames) and $d$ is the hidden size.  
To obtain a fixed-length representation, we apply mean pooling across the token dimension, i.e.,  
\[
\bar{\mathbf{h}}^{(l)}(S) = \frac{1}{T} \sum_{t=1}^T \mathbf{h}^{(l)}_t(S),
\]
which yields a single $d$-dimensional vector per sentence. This simple aggregation strategy has been shown to preserve sentence-level distinctions relevant to grammatical acceptability while avoiding task-specific tuning \citep{pasad2021layer,pasad2023comparative}.  

We then train a logistic regression probe $g$ on the pooled vector $\bar{\mathbf{h}}^{(l)}(S)$ to predict the acceptability label $z \in \{0, 1\}$ for each sentence in a minimal pair, where $z=1$ indicates acceptable.  
All probes are trained using five-fold cross-validation, and evaluated using accuracy as the primary metric.  
By restricting the probe to a linear classifier and keeping the encoder frozen, we ensure that any performance gain reflects information already encoded in the model's representations, rather than introduced through fine-tuning.  
This setup enables a layer-wise analysis of where and how linguistic distinctions—defined by minimal contrasts—are captured in the model.

\paragraph{Probing positional representations.}
To further examine the impact of token aggregation strategy, we additionally evaluate a set of single-token probes, using the hidden state at fixed relative positions (0\%, 25\%, 50\%, 75\%, and 100\%) along the token sequence. This allows us to assess whether specific frames disproportionately encode linguistic features, and to compare the robustness of mean pooling against position-specific representations.

\paragraph{Temporal probing.}
For each sentence we locate the onset of the \emph{critical word}—the element that differentiates the acceptable and unacceptable member of a minimal pair—using WhisperX forced alignment \citep{bain2023whisperx}.  
Around this onset we take a \(\pm1000\text{ ms}\) window and, from the layer that yielded the best mean-pool accuracy, sample individual token embeddings at increasingly fine steps toward the centre.  
Each embedding is fed to a logistic-regression probe, allowing us to trace how well linguistic contrasts are encoded at each time point.

Although the encoders are bidirectional, this analysis still pinpoints \emph{where} in the acoustic stream the model’s internal state carries the strongest grammatical signal.  In continuous speech—where token boundaries are fuzzy—such temporal profiles reveal how structural and conceptual information is distributed over time rather than \emph{when} it is first detected, highlighting the regions the model relies on most for encoding linguistic distinctions.

\paragraph{Selection score.}
% To account for potential biases introduced by model architecture or training setup, we compute a \textit{selection score} that adjusts the measured F1 score of a trained model based on a baseline F1 score obtained from an untrained model. Specifically, the selection score is defined as:
% \[
% S = F1_{\mathrm{trained}} \cdot \left(1 - \frac{\left|F1_{\mathrm{untrained}} - 0.5\right|}{0.5} \right)
% \]
% where \( F1_{\mathrm{trained}} \) denotes the F1 score of the trained classifier, and \( F1_{\mathrm{untrained}} \) is the F1 score obtained from the same model architecture with randomly initialized parameters (i.e., without any training). The intuition behind this metric is that if the untrained model performs close to chance level (i.e., 0.5 in a balanced binary classification setting), the selection score closely reflects the true performance of the trained model. However, if the untrained model already performs significantly above or below chance, the selection score penalizes the trained model accordingly. This helps prioritize examples or tasks where training contributes meaningfully beyond architectural bias.

To reflect asymmetric effects of architectural bias, we define a selection score that adjusts the trained model's accuracy based on whether the untrained model performs above or below chance level:
\[
Selec = Acc_{\mathrm{trained}} \cdot \left(1 + \frac{0.5 - Acc_{\mathrm{untrained}}}{0.5} \right)
\]
% This formulation rewards stronger gains when the untrained model performs below chance, and penalizes when the untrained model performs above chance. It provides an interpretable metric of model improvement relative to the architectural baseline.
This formulation rewards stronger gains when the untrained model performs below chance, and penalizes when it performs above chance. For speech models, although untrained networks contain no learned parameters, the input spectrograms often carry meaningful structure that can lead to above-chance performance. The selection score thus quantifies the improvement achieved by training, relative to the representational bias inherent in both the model architecture and the input features.

\paragraph{Confidence score.}
% The evidence score is computed as follows: First, we define the set of correctly classified samples as 
% \[
% \mathcal{C} = \{ i \mid y_{\mathrm{true},i} = y_{\mathrm{pred},i} \}
% \]
% where \( y_{\mathrm{true}} \) represents the ground truth labels and \( y_{\mathrm{pred}} \) represents the predicted labels. The correct classification rate (accuracy) is given by
% \[
% A = \frac{|\mathcal{C}|}{n}
% \]
% where \( n \) is the total number of test samples. Next, for each correctly classified sample, we extract the maximum predicted probability across all classes and compute the mean confidence score for these samples as
% \[
% E_{\mathrm{correct}} = \frac{1}{|\mathcal{C}|} \sum_{i \in \mathcal{C}} \max_j P_{i,j}
% \]
% where \( P_{i,j} \) is the probability of sample \( i \) belonging to class \( j \). Finally, the overall evidence score is calculated by incorporating both accuracy and confidence:
% \[
% \mathrm{Evidence} = E_{\mathrm{correct}} \times A
% \]
% This formulation ensures that the evidence score reflects both the model's classification confidence for correctly predicted samples and its overall classification accuracy.
The confidence score is computed as follows: First, we define the set of correctly classified samples as \( \mathcal{A} = \{ i \mid y_{\mathrm{true},i} = y_{\mathrm{pred},i} \} \), where \( y_{\mathrm{true}} \) represents the ground truth labels and \( y_{\mathrm{pred}} \) represents the predicted labels. Next, for each correctly classified sample, we extract the maximum predicted probability across all classes and compute the mean confidence score for these samples as 
\[
Conf = (1 / |\mathcal{A}|) \sum_{i \in \mathcal{A}} \max_j P_{i,j} 
\],
where \( P_{i,j} \) is the probability of sample \( i \) belonging to class \( j \). 
Measuring the confidence score allows us to assess how certain the model is when it makes correct predictions, offering insight into the calibration of its probabilistic outputs. A well-calibrated model should not only predict correctly but also assign high confidence to those correct predictions.

% The confidence score is computed as follows: First, we define the set of correctly classified samples as \( \mathcal{C} = \{ i \mid y_{\mathrm{true},i} = y_{\mathrm{pred},i} \} \), where \( y_{\mathrm{true}} \) represents the ground truth labels and \( y_{\mathrm{pred}} \) represents the predicted labels. The correct classification rate (accuracy) is given by \( A = |\mathcal{C}| / n \), where \( n \) is the total number of test samples. Next, for each correctly classified sample, we extract the maximum predicted probability across all classes and compute the mean confidence score for these samples as \( E_{\mathrm{correct}} = (1 / |\mathcal{C}|) \sum_{i \in \mathcal{C}} \max_j P_{i,j} \), where \( P_{i,j} \) is the probability of sample \( i \) belonging to class \( j \). Finally, the overall confidence score is calculated as \( \mathrm{Confidence} = E_{\mathrm{correct}} \times A \), ensuring that it reflects both the model's classification confidence for correctly predicted samples and its overall classification accuracy.

\section{Results} 
\subsection{Trained representations cluster by linguistic category and task}
\begin{figure}[h]
    \centering
    \includegraphics[width=0.48\textwidth]{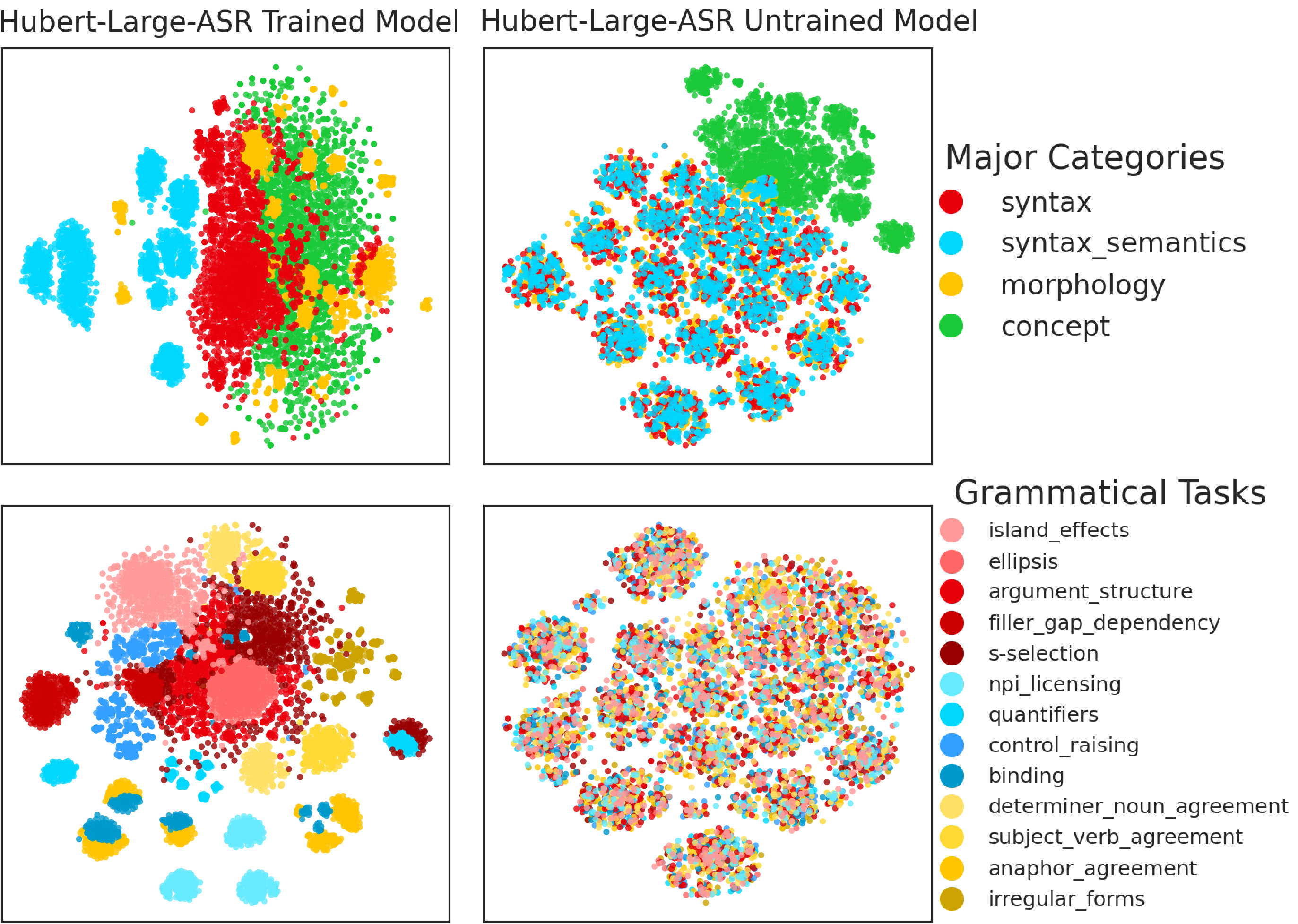}
    \caption{\textbf{t-SNE visualisation of minimal–pair \(\Delta\)-embeddings in \textsc{HuBERT-Large-ASR} (layer~18).}
For each sentence we compute the difference vector \(\Delta\mathbf h = \bar{\mathbf h}(S^{+}) - \bar{\mathbf h}(S^{-})\) between the acceptable and unacceptable member of a minimal pair, then project all \(\Delta\mathbf h\) to two dimensions with t-SNE (perplexity = 50).  
\emph{Top-left:} trained model coloured by major category (Syntax~\textcolor{red}{\textbullet}, Syntax–Semantics Interface~\textcolor{cyan}{\textbullet}, Morphology~\textcolor{orange}{\textbullet}, Concept~\textcolor{green}{\textbullet}).  
\emph{Top-right:} randomly initialised model.  
\emph{Bottom-left:} trained model coloured by twelve sub-tasks within the first three categories.  
\emph{Bottom-right:} corresponding untrained control.  
% The trained encoder forms clear clusters for Syntax, Interface, and Morphology—and for their finer sub-tasks—whereas Conceptual contrasts and all untrained projections remain largely intermixed, confirming that linguistic structure emerges only after self-supervised learning.
}

\label{fig:tsne}
\end{figure}

The trained encoder forms well-defined clusters that are absent in the randomly-initialised network, indicating that self-supervised pre-training imbues the model with embeddings that systematically separate grammatical contrasts. Among the four major categories, Syntax, Syntax–Semantics interface, and Morphology occupy distinct regions, whereas Conceptual contrasts remain intermixed—mirroring the weaker probing accuracies reported earlier. Finer structure emerges in the bottom-left panel: sub-tasks such as island effects, ellipsis, and agreement each form coherent sub-clusters, showing that the model encodes nuanced distinctions within broader grammatical families. In the untrained baseline (bottom right) these patterns collapse into a homogeneous cloud, confirming that the observed structure arises from linguistic learning rather than architectural priors alone.

\begin{figure}[h]
    \centering
    \includegraphics[width=1.0\linewidth]{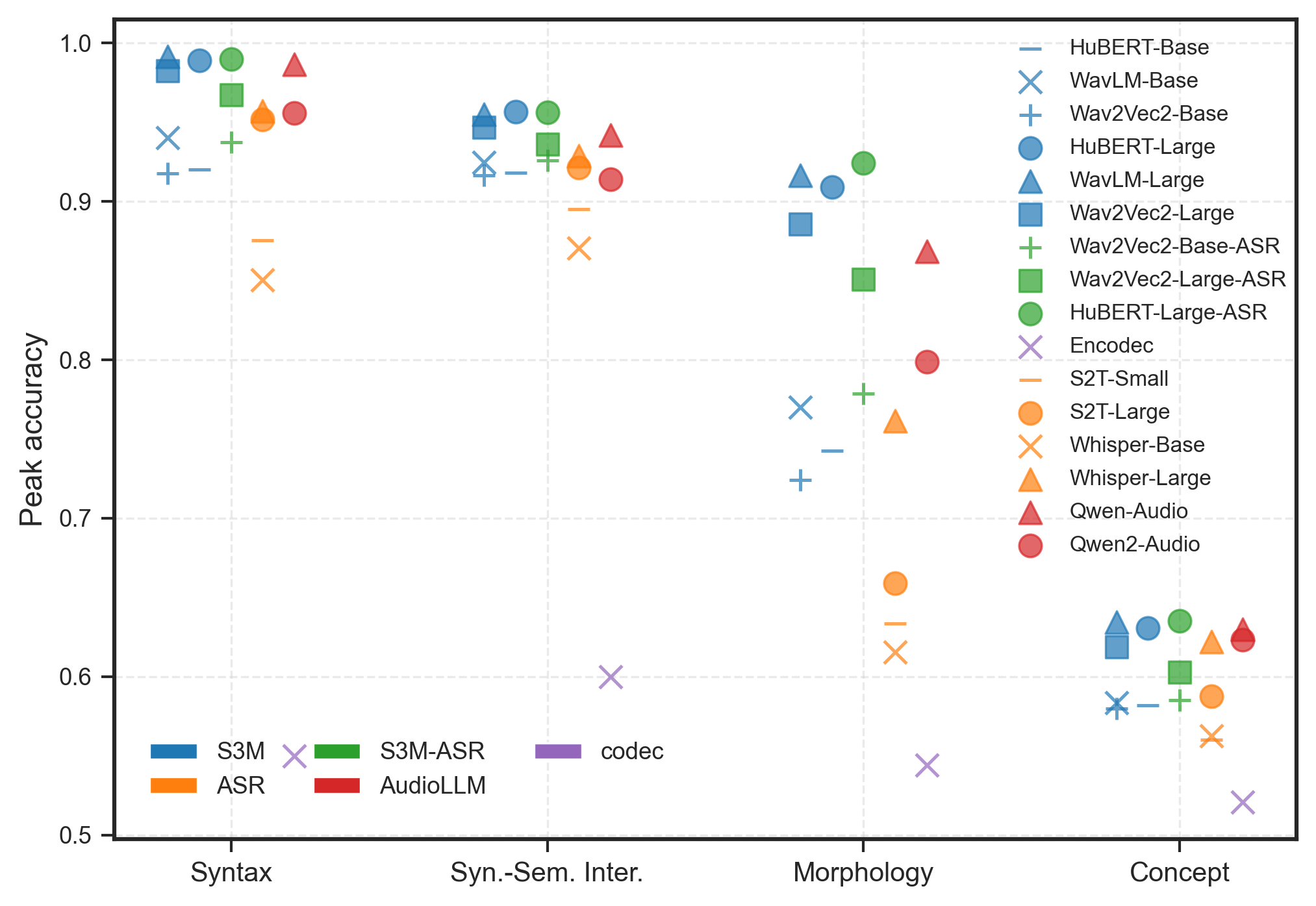}
    \caption{\textbf{Peak minimal-pair probing accuracy across linguistic levels.} For each model, we report the highest sentence-level accuracy attained by any layer on syntax, syntax–semantics interface, morphology, and concept tasks.}
    \label{fig:peak_f1}
\end{figure}

\begin{figure*}[h]
    \centering
    \includegraphics[width=1.0\linewidth]{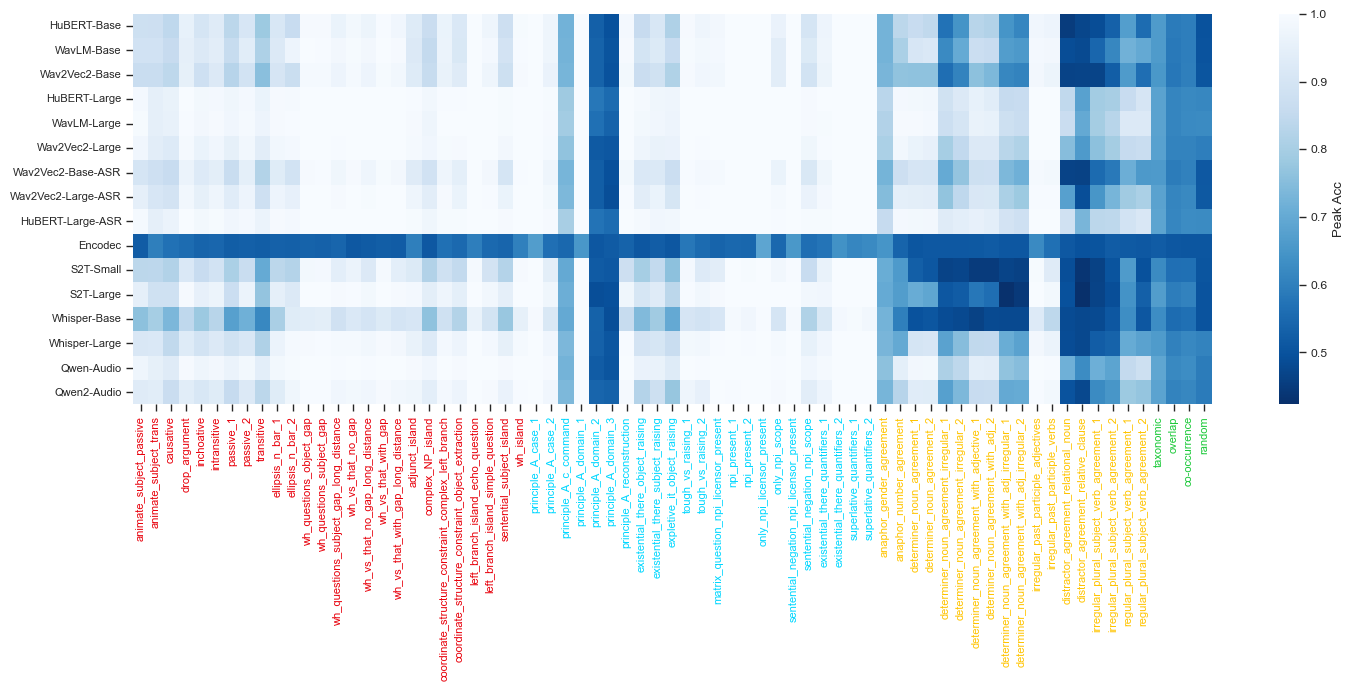}
    \caption{\textbf{Peak accuracy for every individual minimal-pair task across models.} Rows list the 16 speech encoders; columns list the 67 grammatical tasks: syntax (red), syntax-semantics interface (blue), morphology (yellow) and 4 conceptual tasks (green).}
    \label{fig:peak_results}
\end{figure*}

\begin{figure*}[h]
    \centering
    \includegraphics[width=1.0\linewidth]{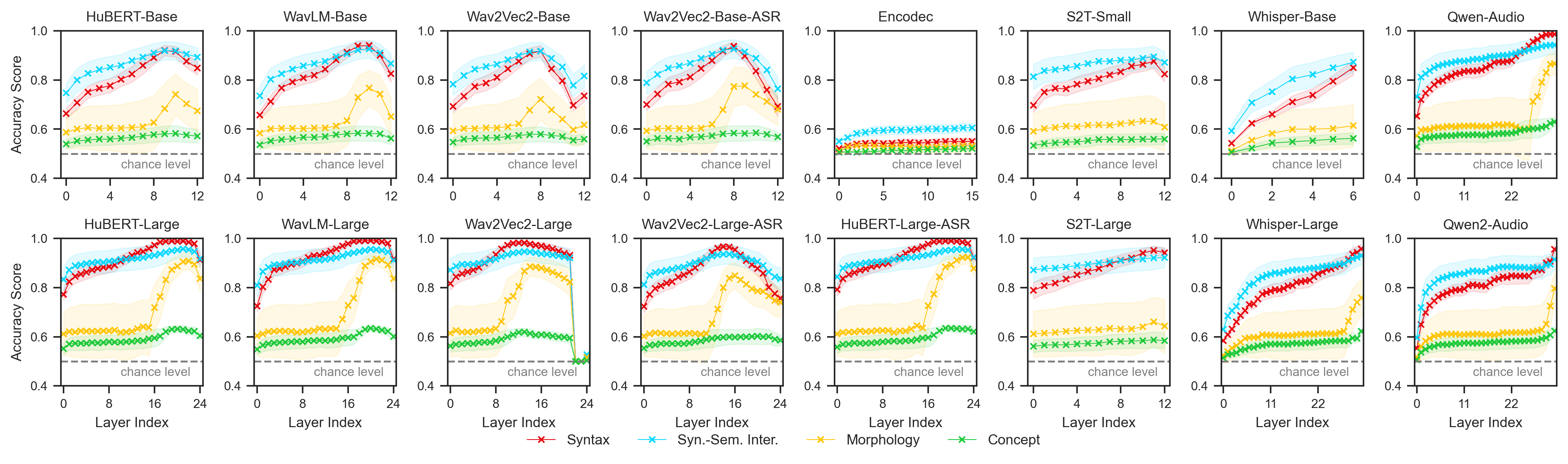}
    \caption{\textbf{Layer-wise trajectory of linguistic accuracy for each speech model.} For every encoder we plot sentence-level probing accuracy (y-axis) as a function of Transformer depth (layer index, x-axis). Shaded bands denote ±1 standard error over five cross-validation folds.}
    \label{fig:layerwise_f1}
\end{figure*}

% From our results, we identify several key findings on how linguistic information is encoded within speech models: form over meaning, text-free linguistic learning, and hierarchical transformer encoding.

\subsection{Speech models encode form better than meaning.}
We first demonstrate a clear disparity in how effectively speech models capture linguistic ``form'' versus ``meaning''. As evident from Figure \ref{fig:peak_f1},  grammatical minimal pairs (including syntax, syn.-sem. interface and morphology) are consistently distinguished better with a higher Acc score than conceptual pairs for all models. This performance difference in the structured nature of syntactic and morphological rules makes them more readily learnable from acoustic signals alone, while the abstraction required for conceptual understanding poses greater challenges. The robust encoding of linguistic form indicates that speech models implicitly acquire rule-based structural regularities even in the absence of explicit textual guidance.

\subsection{Speech transformer layers show hierarchical linguistic features.}
% We further observe that speech models encode linguistic features hierarchically across transformer layers. First, as anticipated, larger models (Large vs. Base) with more layers achieve stronger predictive performance for linguistic ambiguities. Within each model, the capability to resolve syntactic, morphological, and conceptual distinctions generally increases monotonically from shallow to deeper layers. However, peak performance occurs slightly before the final layers for self-supervised speech models wav2vec 2.0, HuBERT, and WavLM. This phenomenon arises because the last few layers are predominantly optimized toward their specific self-supervised tasks, making earlier layers more effective at general linguistic representation. This hierarchical encoding pattern emphasizes a universal strategy in transformer language models, progressively abstracting simpler acoustic signals into abstract linguistic features. 
As shown in Figure~\ref{fig:layerwise_f1}, we observe a clear hierarchical emergence of linguistic representations across transformer layers in speech models. Syntactic and syntax–semantics interface features are learned earliest, showing robust gains beginning from the lower-middle layers and typically peaking around the upper-middle layers. Morphological features, in contrast, emerge much later—often only becoming linearly decodable in the final few layers. Conceptual features exhibit a different trend: while they become distinguishable earlier than morphology, their decodability remains consistently weak across all layers.

This pattern suggests that structural distinctions like syntax can be encoded relatively early from local phonological and prosodic cues, while morphological information requires more global context and fine-grained form recognition—hence emerging later. Conceptual meaning, on the other hand, likely requires a fundamentally different abstraction capacity that is not easily captured by current speech models, even at their deepest layers.

Interestingly, while larger models (e.g., Wav2Vec2-Large, HuBERT-Large, WavLM-Large) consistently outperform their base counterparts, we find that performance for most linguistic categories saturates before the final few layers. This supports the view that uppermost layers in self-supervised models are often specialized for pretraining objectives (e.g., contrastive or masked prediction), and that mid-to-upper layers more faithfully encode linguistic structure. These findings highlight a progressive and modular abstraction process in speech transformers, wherein linguistic features emerge at different depths according to their complexity and signal grounding.

% \subsection{Self-supervised speech models can learn linguistic knowledge without text.}
% Self-supervised speech models (S3Ms), trained without textual supervision, exhibit remarkable capabilities in acquiring linguistic knowledge solely from audio signals, as illustrated in Figure \ref{fig:peak_f1}. Specifically, wav2vec 2.0, HuBERT, and WavLM achieve impressive performance across linguistic levels, with Acc scores above 90\% on syntactic and syn.-sem. interface minimal pairs, between 80\% and 90\% on morphology tasks, and around 60\% on conceptual distinctions. Interestingly, their overall performance surpasses ASR models explicitly trained with textual supervision. One plausible explanation is that ASR models may overfit to transcription objectives, thus mainly recognizing linguistic patterns as they appear to, whereas S3Ms develop deeper and more robust linguistic representations. The strong representations developed by S3Ms are mainly due to their training objectives, such as wav2vec 2.0's contrastive learning and HuBERT and WavLM's masked audio prediction, both promoting the model's learning context beyond the locality of spectrograms. These results highlight the promise of self-supervised acoustic training paradigms for generalizable linguistic learning from unlabeled audio data.

\begin{figure*}[t]
    \centering
    \includegraphics[width=1.0\linewidth]{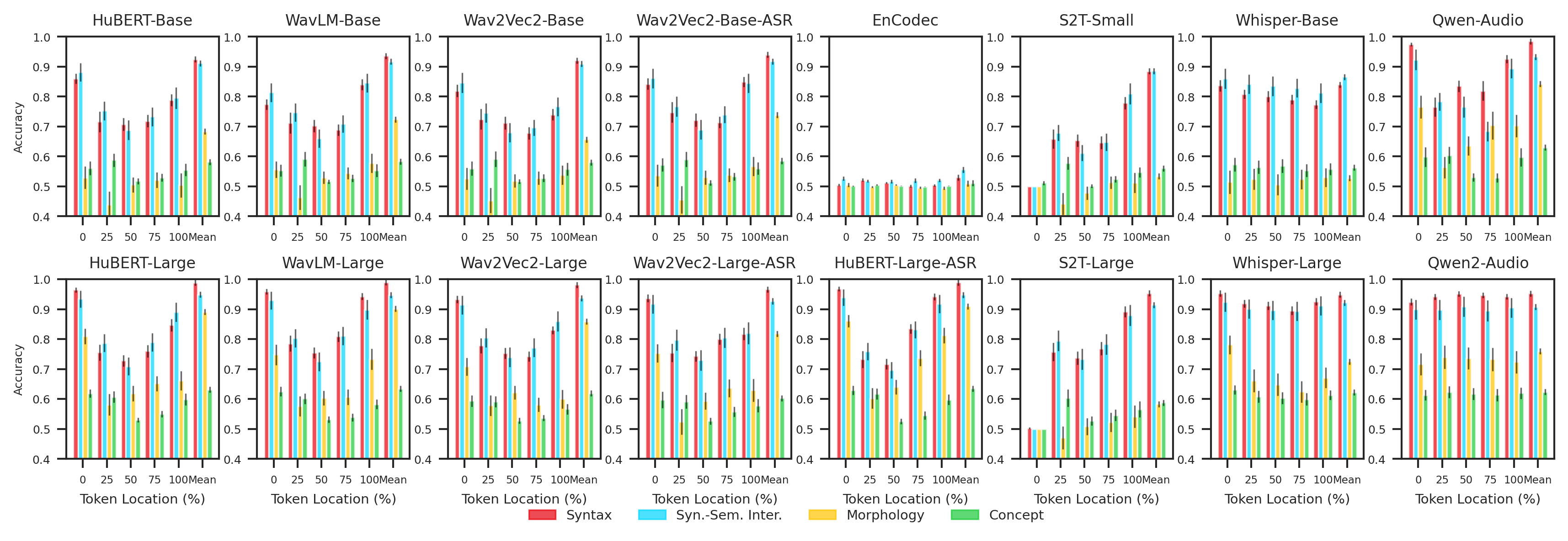}
    \caption{\textbf{Comparing single-token versus mean pooling across speech models.} We evaluate probing accuracy when sentence embeddings are extracted from a single token at five relative positions (0\%, 25\%, 50\%, 75\%, 100\%) or via mean pooling across all tokens ("Mean"). Accuracy is measured on the layer that gave the best mean-pool score.}
    \label{fig:token_positions}
\end{figure*}

\subsection{Self-supervised speech models can learn linguistic knowledge without text}
Self-supervised speech models (S3Ms), trained without textual supervision, exhibit remarkable capabilities in acquiring linguistic knowledge solely from audio signals, as illustrated in Figure~\ref{fig:peak_f1} and \ref{fig:layerwise_f1}. Specifically, wav2vec~2.0, HuBERT, and WavLM achieve impressive performance across linguistic levels, with accuracy scores exceeding 90\% on syntactic and syntax–semantics interface tasks, reaching 80–90\% on morphology, and around 60\% on conceptual distinctions. Surprisingly, these models outperform ASR encoders trained with explicit text supervision.

One plausible explanation is that ASR models, while grounded in linguistic output, are optimized primarily for surface-level transcription. This objective may bias them toward segmental and lexical patterns that are useful for decoding text but less aligned with abstract grammatical distinctions. In contrast, S3Ms benefit from pretraining objectives such as contrastive learning (wav2vec~2.0) and masked prediction (HuBERT, WavLM), which promote contextual abstraction and robustness beyond frame-level cues. These results highlight the promise of self-supervised acoustic pretraining as a pathway toward generalizable linguistic learning from raw audio.

\subsection{Architectural and training differences shape linguistic depth}
While ASR and AudioLLM models share similar encoder backbones (e.g., Transformer-based encoders with convolutional frontends), their training paradigms differ substantially. AudioLLMs (e.g., Qwen-Audio) are trained under multimodal generative objectives within large-scale LLM frameworks. As shown in Figure~\ref{fig:peak_f1}, their performance on syntactic and morphological tasks consistently exceeds that of ASR models, suggesting that joint language–audio modeling drives richer linguistic abstraction even in the encoder alone. This indicates that the nature of the downstream supervision—not just the model capacity—plays a critical role in shaping linguistic competence.

In addition, we observe a distinct architectural signature: across all S3M models, performance drops sharply in the final few layers. This effect is notably absent in ASR and AudioLLM encoders, where accuracy continues to rise or plateaus in the top layers. The late-layer drop in S3Ms likely reflects a shift in focus toward pretraining-specific representations—such as contrastive codebook prediction or masked target reconstruction—rather than general-purpose linguistic encoding. This divergence underscores a fundamental difference: while S3Ms produce general linguistic features as a byproduct of self-supervised objectives, ASR and LLM-based models are tuned explicitly for language-level outputs, preserving linguistic structure deeper into the network.

\begin{figure*}[h]
    \centering
    \includegraphics[width=1.0\textwidth]{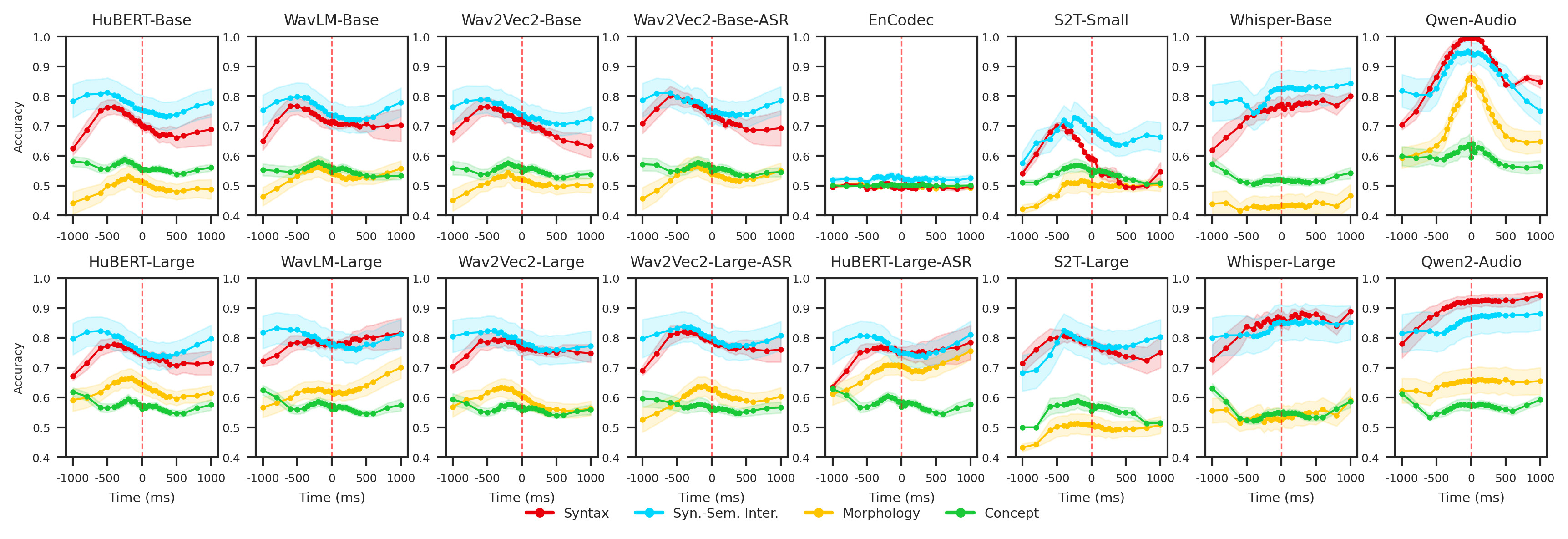}
    \caption{\textbf{Temporal dynamics of linguistic information around the critical word.}
For each model we plot probing accuracy as a function of time relative to the onset of the critical word (vertical dashed line = 0 ms). Accuracy is measured on the layer that gave the best mean-pool score.}
  \label{fig:temporal-dynamic-probing}
\end{figure*}

\subsection{Mean pooling outperforms single-token alternatives in speech models}

A common practice in large language model research is to extract sentence-level representations using the final token embedding—particularly in autoregressive (unidirectional) architectures. To evaluate whether similar heuristics apply to speech models, we compare mean pooling with single-token probing at five fixed relative positions (0\%, 25\%, 50\%, 75\%, and 100\%) along the token sequence. As shown in Figure~\ref{fig:token_positions}, mean pooling consistently yields higher probing accuracy across all linguistic categories and models, confirming its robustness for speech-based sentence embedding.

Surprisingly, although most speech models are based on bidirectional Transformers and thus in principle have symmetric access to context, we consistently find that the first token (0\%) encodes more linguistic information than the final token (100\%). This asymmetry is especially evident in models like HuBERT, Wav2Vec2, and WavLM. One possible explanation is that the first token may serve as an implicit global aggregator, especially in architectures without explicit [CLS] tokens. During training, early tokens might become more sensitive to contextual summaries or more stable across time, while final tokens are often affected by alignment noise or signal boundary effects. This positional imbalance suggests that even in bidirectional encoders, representational asymmetry can emerge as a byproduct of training dynamics and audio processing pipelines. While mid-sequence tokens (e.g., 25\%, 50\%) sometimes approach mean pooling performance, no single-token baseline reliably matches it, reinforcing the importance of full-sentence integration in capturing hierarchical linguistic features in speech.

\subsection{Temporal asymmetry despite bidirectional context}
% \footnote{More results regarding selection score and confidence score could be found in Appendix \ref{sec:analysis}.}

As Figure~\ref{fig:temporal-dynamic-probing} shows, although all encoders are bidirectional, their temporal profiles are strikingly \emph{asymmetric}. In the three self-supervised families (HuBERT, WavLM, Wav2Vec2) and in the S2T ASR models, syntactic and interface accuracy peaks -600 to -500 ms \textbf{before} the critical word begins, then decays after onset. A plausible explanation is that these models learn to anticipate upcoming phone sequences from co-articulatory cues: masked-prediction and CTC-style contrastive objectives reward early integration of left-context information, causing pre-onset tokens to act as forward predictors.

By contrast, Whisper and Qwen-Audio—whose encoders are trained inside multimodal generative frameworks—display a flatter, gradually rising curve that culminates around or shortly after onset. These models appear to accumulate evidence more evenly across the word, perhaps because their objectives (sequence-to-sequence transcription or audio-text alignment) emphasise complete lexical identification rather than early prediction.

Morphological information emerges later and more slightly symmetrically, reflecting the need for full word forms, while conceptual distinctions remain weak throughout the window. EnCodec stays at chance, confirming the absence of linguistically relevant structure in a purely compression-oriented codec. Together, these results show that speech encoders encode grammatical distinctions at different \emph{timescales}: S3M and S2T models exploit anticipatory acoustic cues, whereas AudioLLM encoders rely on longer integration windows—revealing how training objectives shape not only \emph{where} but also \emph{when} linguistic knowledge is realised in the representation.

\section{Conclusion}
By first introducing minimal pairs into speech probing, huoshow that contemporary speech transformers internalise a structured linguistic hierarchy. 
% Syntax becomes reliably accessible first, followed by interface cues and, much later, morphology, whereas conceptual knowledge is present but less robust. 
% Self-supervised objectives are surprisingly effective, delivering representations that outstrip those of text-supervised ASR systems and approach the quality of multimodal AudioLLMs. 
% Layer-wise and temporal analyses expose objective-specific signatures: S3Ms trade late-layer linguistic fidelity for pretraining targets, and only models optimised for language-level outputs preserve grammar to the top. 
Our results suggest 
% that future speech models should combine the context-rich learning of S3Ms with objectives that explicitly ground meaning, and
that evaluation of spoken-language understanding should move beyond coarse benchmarks to fine-grained, temporally resolved diagnostics.
% \section*{Acknowledgments}

% \clearpage
\section*{Limitation}
\subsection*{Language and audio diversity}
All experiments use English sentences rendered by a single, studio-quality TTS voice. Consequently, the reported hierarchy may be specific to English morpho-syntax and to clean, single-speaker recordings. Languages with richer inflection (e.g., Turkish) or tonal contrasts (e.g., Mandarin), as well as spontaneous, multi-speaker, or noisy speech, might yield very different layer-wise and temporal patterns.

\subsection*{Scope of the linguistic tests}
Our 116,300 minimal pairs target core morpho-syntactic phenomena but exclude discourse-level skills such as anaphora, presupposition, or pragmatic implicature. The strong form–meaning gap we observe therefore says little about how models handle context-dependent or world-knowledge-rich interpretations that dominate real conversation.

% \subsection{Linear probing as a diagnostic}
% We rely on logistic-regression probes to avoid inflating scores, yet linear separability is only a lower bound on what the network encodes. Non-linear decoders could reveal hidden structure, while intervention studies would be needed to show that the probed features are actually \emph{used} by the model during inference.

\subsection*{Bidirectionality and temporal claims}
The encoders are bidirectional, so pre-onset peaks do not prove real-time anticipation; they merely locate where the internal state most strongly distinguishes the contrast. True causal timing would require strictly streaming models or masking future context during probing.

\subsection*{Model and parameter range}
We analyse 16 open-weight encoders while larger proprietary AudioLLMs and causal streaming ASR systems are absent, and scaling laws above the tested range remain unexplored.

\subsection*{Absence of speech-specific minimal pairs}
Our benchmark manipulates linguistic content while holding the acoustic signal constant, but it does not include \emph{phonetic} or \emph{prosodic} minimal pairs—e.g.\ vowel length contrasts, stress-shift alternations, or intonation-driven meaning changes. As a result, we cannot assess whether the same hierarchical pattern extends to speech-specific cues such as co-articulation, pitch accent, or rhythm, nor can we determine how lexical and segmental information interact in the encoder. A companion suite of acoustically controlled minimal pairs would be required to map the boundary between linguistic and purely phonetic knowledge in self-supervised speech models.

\subsection*{No neuron-level dissection}
All conclusions are drawn from layer-global probes; we do not inspect individual neurons or sparse subnetworks that might carry specialised features. Prior work in text LLMs has shown that single units can act as high-precision detectors for concepts like negation or syntax islands. Without analogous neuron-level analyses, we cannot say whether speech encoders store linguistic knowledge in distributed patterns, in a handful of specialised channels, or in attention heads. Such fine-grained dissection could reveal bottlenecks, redundancies, or pathways amenable to targeted editing.

\subsection*{Unanalysed attention structure}
Transformer attention maps provide a window into how models route information, yet our study treats them as a black box. We do not test whether attention heads focus on the critical word, respect syntactic dependencies, or exhibit long-range patterns akin to filler–gap tracking. Consequently, we cannot link the probing results to specific computational mechanisms inside the model. Future work combining attention diagnostics with our temporal and layer-wise probes could show \emph{how}—not just \emph{where}—speech transformers build the observed hierarchy of linguistic representations.
\clearpage
\bibliography{references}

\clearpage
\appendix
\section{Related Work}
% Research on the probing of neural speech models has evolved along three main lines: (i) analyses of low-level acoustic and phonetic representations, (ii) investigations of higher-level contextual linguistic features, and (iii) probing studies that connect learned representations with downstream performance and integrated systems.
\subsection{Low-Level Acoustic and Phonetic Probing} Early investigations focused mainly on how neural models capture fundamental acoustic signals and phonetic details. The analyses of end-to-end ASR systems by \citep{belinkov2017analyzing, belinkov2019analyzing} revealed that phonetic and graphemic information is distributed in a layer-dependent manner, highlighting the role of early layers in the capture of fine-grained details. Extending these insights to self-supervised settings, \citep{ma2021probing} and \citep{de2022probing} demonstrated that models such as Contrastive Predictive Coding effectively capture phoneme-level distinctions. Researchers have employed techniques such as canonical correlation analysis and mutual information measures to trace the evolution of these representations. These methods reveal that early layers are particularly sensitive to subtle cues—including accent variations and phonetic nuances—as shown by studies such as \citep{prasad2020accents, raymondaud2024probing}. Comparative analyses of transformer architectures further illustrate diverse learning patterns for low-level acoustic features \citep{chung2021similarity, yang2020understanding, shah2021all}, underscoring the foundational role of these representations in speech processing.

\subsection{Linguistic Feature Probing} 
% Beyond basic acoustics, a growing body of work has focused on the encoding of semantic and syntactic information within speech models. Benchmark frameworks like SpeechGLUE \citep{ashihara2023speechglue} have been instrumental in systematically evaluating linguistic knowledge, revealing that while speech models capture cues such as word identity and boundaries, they often prioritize phonetic similarity over deeper semantic content. Supporting this observation, \citep{pasad2024self} and \citep{choi2024self} illustrate that the distribution of linguistic information is strongly influenced by the pre-training objective and model scale. Detailed layer-wise analyses \citep{pasad2021layer, pasad2023comparative} indicate that deeper layers tend to encode more abstract syntactic and contextual features. Minimal pair probing for syntactic cues \citep{shen2023wave} and auditory grounding studies \citep{ngo2024language} further reinforce the notion that higher-level linguistic properties gradually emerge across the network. Additionally, probing of deep speaker embeddings has shown that some layers effectively disentangle linguistic content from speaker-specific information \citep{zhao2022probing, ashihara2024self}, offering nuanced insights into the multifaceted nature of speech representations.
Beyond basic acoustic and lexical feature analyses, several studies have probed deeper linguistic encodings in speech models. 
% Early work focused on phonetic and phonological representations, showing that self-supervised speech models capture fine-grained phone identity and boundary information \citep{belinkov2017analyzing, ma2021probing, wells2022phonetic}.
Layer-wise investigations revealed that phonology and morphology emerge predominantly in lower to mid layers \citep{pasad2021layer, pasad2024self}, while speaker-identity probes demonstrated partial disentanglement of speaker-specific traits from linguistic content \citep{williams2019disentangling, raj2019probing}. Word-level semantic evaluations—most notably via SpeechGLUE—have assessed static lexical semantics, indicating that many models still prioritize phonetic similarity over deeper meaning \citep{ashihara2023speechglue}. Shallow syntactic metrics (e.g., part-of-speech tags, dependency tree depth) have also been applied, suggesting that basic syntactic cues are learnable but without distinguishing long-range dependencies \citep{martin2023probing, pasad2021layer, shen2023wave}.

\subsection{Probing in Downstream Tasks and Integrated Systems} The practical utility of speech representations is underscored by their impact on downstream applications, where the quality of internal representations directly influences system performance. Early efforts in this domain include the unified speech language pre-training framework proposed by \citep{qian2021speech}, which integrates an ASR encoder with a language model to improve spoken language understanding. Subsequent studies by \citep{chang2024exploring} and \citep{abdullah2023information} have explored the use of discrete speech units to compress input sequences while maintaining robust performance in ASR, translation, and comprehension tasks. Benchmark studies \citep{zaiem2025speech} further highlight that the architectural design of probing heads can critically affect model rankings and task outcomes. Statistical analyzes \citep{ollerenshaw2022insights} linking layer-wise properties to recognition accuracy underscore the direct relationship between internal representations and practical performance. More recent evaluations of emerging SpeechLLMs \citep{wu2024just} point to challenges such as limited speaker awareness, even as these models demonstrate impressive capabilities in context-based dialogue tasks. This line of inquiry emphasizes that the effectiveness of speech models in real-world applications depends on a delicate balance between capturing low-level signals and encoding higher-level linguistic and contextual features.

\section{Models Details}
\label{sec:models}
\subsection{Probed Models}
\paragraph{Self-Supervised Speech Models (S3Ms)}
Modern self-supervised learning paradigms learn robust speech representations through pre-training objectives that do not require transcribed data:
\begin{itemize}
    \item \textbf{HuBERT} \citep{hsu2021hubert}: HuBERT employs an iterative clustering procedure combined with a masked prediction objective to jointly model acoustic and linguistic characteristics. We analyze both the \texttt{facebook/hubert-large-ll60k} and the \texttt{facebook/hubert-base-ls960} variants in our experiments.
    \item \textbf{wav2vec 2.0} \citep{baevski2020wav2vec}: This model employs latent space masking with contrastive learning, where representations are learned jointly with a quantized codebook. Our study includes the \texttt{facebook/wav2vec2-base} and \texttt{facebook/wav2vec2-large-lv60} models.
    \item \textbf{WavLM} \citep{chen2022wavlm}: WavLM extends self-supervised learning by integrating a denoising objective with masked speech prediction, thereby enhancing its robustness across a variety of speech tasks. We consider both the \texttt{microsoft/wavlm-base-plus} and the \texttt{microsoft/wavlm-large} variants.
\end{itemize}

\paragraph{Automatic Speech Recognition (ASR) Models}
ASR models typically achieve high transcription accuracy while embedding additional linguistic biases. Our experiments probe only the encoder components of these models, extracting representations before the decoder stage:

\begin{itemize}
    \item \textbf{Speech-to-Text Models} \citep{wang-etal-2020-fairseq}: The S2T model provides transformer-based architectures for end-to-end speech recognition and translation, and integrates acoustic modeling with language model fusion. We study both the \texttt{facebook/s2t-small-librispeech-asr} and \texttt{facebook/s2t-large-librispeech-asr} configurations to assess how model complexity influences downstream performance.
    \item \textbf{Whisper} \citep{radford2023robust}: Trained in a weakly supervised extensive corpus, Whisper demonstrates robust performance in various speech conditions. Our experiments involve the \texttt{openai/whisper-base} and \texttt{openai/whisper-large} variants.
\end{itemize}

\paragraph{Auditory Large Language Models (AudioLLMs)}
AudioLLMs fuse the strengths of acoustic modeling with the advanced reasoning capabilities of large-language models. Similar to our ASR experiments, we probe only the audio encoder components of these models, allowing direct comparison with other speech encoders:
\begin{itemize}
    \item \textbf{Qwen Audio} \citep{chu2023qwen}: A multimodal large language model that accepts speech, natural sounds, music and text, producing textual outputs. It employs a multi-task learning framework that enables knowledge sharing across more than 30 tasks while avoiding one-to-many interference.    
    \item \textbf{Qwen2 Audio} \citep{chu2024qwen2}: An evolution of the original Qwen Audio. It integrates a dedicated audio front-end (initialized from Whisper) with the Qwen-7B language model backbone. It simplifies pre-training by using natural language prompts rather than hierarchical tags. 
    % \item \textbf{Salmonn} \citep{tang2023salmonn}: By integrating pre-trained speech and audio encoders with a text-based large language model, Salmonn facilitates a wide range of tasks including ASR, translation, and audio captioning. Its design emphasizes cross-modal reasoning, yielding emergent capabilities such as audio-based storytelling.
\end{itemize}

\paragraph{EnCodec \cite{defossez2022high}}
Encodec is a neural audio codec that provides real-time, high-fidelity compression through a streaming encoder-decoder architecture with quantized latent space. It employs a multiscale spectrogram adversary to reduce artifacts and a novel loss balancer mechanism. Our experiments utilize the \texttt{facebook/encodec\_24khz} model variant. For probing EnCodec, we extract representations from each quantizer by passing audio through the encoder and accessing the discrete codes at each quantizer. This allows us to analyze how linguistic information is encoded at different levels of the quantization hierarchy.

\subsection{Text2Speech Generation}
We generated speech minimal pairs by synthesizing speech from minimal pairs in text using the Kokoro-82M text-to-speech (TTS) model. Kokoro is a state-of-the-art human-level TTS model based on the StyleTTS 2 \cite{li2023styletts} architecture and ISTFTNet \cite{kaneko2022istftnet} vocoder. The output 24 KHz speech waveforms were subsequently resampled to 16 kHz, aligning with the sampling rate of speech models. Additionally, we manually reviewed synthesized speech samples to ensure they contained no acoustic glitches and that the incorrect sentences were not unintentionally auto-corrected.
\begin{figure*}[h]
    \centering
    \includegraphics[width=1.0\linewidth]{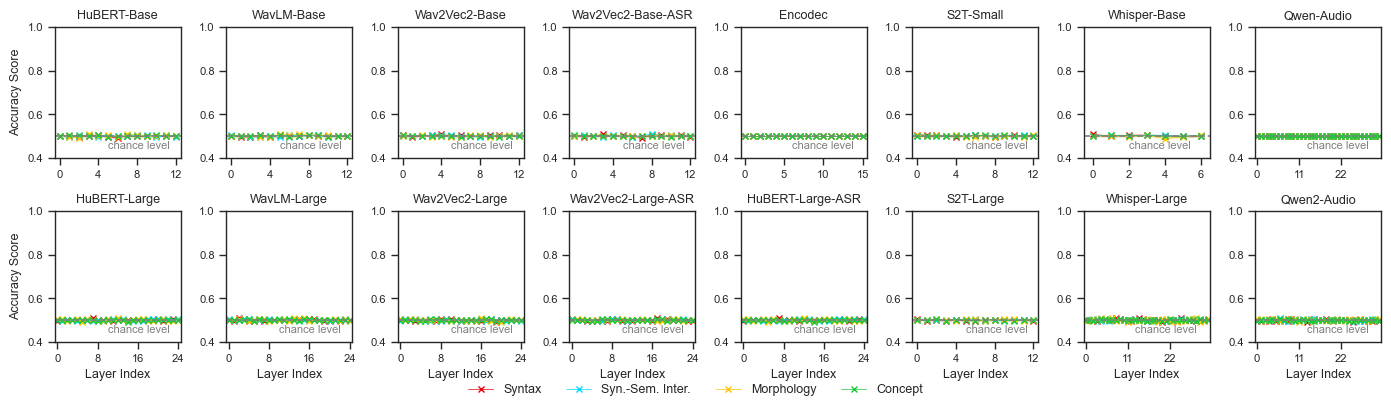}
    \caption{\textbf{Control experiment with random embeddings.}  
To verify that probe accuracy reflects genuine linguistic encoding rather than dataset biases, we conduct an ablation in which each speech input is replaced with a random embedding sampled from a Gaussian distribution matching the original mean and standard deviation of that model’s sentence embeddings.}

    \label{fig:layerwise_random}
\end{figure*}

% \section{Ablation tests} 
% As shown Figure~\ref{fig:layerwise_random}, ablation tests that shuffle the acceptability labels of the pairs reduce probe accuracy to chance level, implying that information useful for the task is concentrated precisely in the manipulated position—strong evidence for the diagnostic precision of the minimal-pair methodology.

\section{Ablation: probing with matched random embeddings.}  
To ensure that our probing results reflect meaningful structure in model representations—rather than spurious correlations exploited by the linear classifier—we conduct a control experiment using randomly generated embeddings.

Specifically, for each sentence, we generate a random vector sampled from a Gaussian distribution that matches the mean and standard deviation of the original sentence embeddings from the best-performing layer of the model. To preserve the pairing structure of the data, we assign the same random vector to both minimal-pair variants if they are based on the same audio clip. These random embeddings are then passed through the same probing pipeline, including five-fold logistic regression and accuracy evaluation.

Unlike the \emph{randomly initialized model} condition—which still processes structured spectrograms through a frozen encoder—this ablation entirely removes content information by bypassing the encoder and directly replacing model representations with noise. As shown in Figure~\ref{fig:layerwise_random}, probing accuracy drops to chance for all models and linguistic categories, confirming that the original signal arose from linguistic encoding in the representations—not from classifier bias or residual acoustic confounds. This further validates the diagnostic precision of our minimal-pair probing setup.

\section{More Layer-wise Analysis}
\label{sec:analysis}
\begin{figure*}[h]
    \centering
    \includegraphics[width=1.0\linewidth]{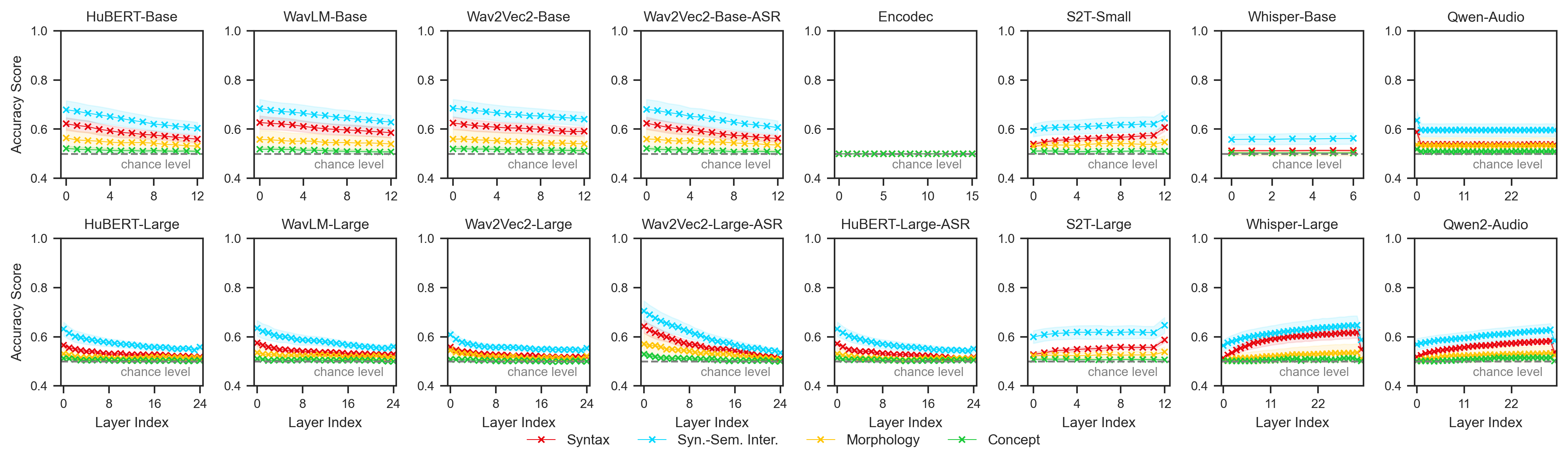}
    \caption{Layer-wise accuracy of randomly initialized (untrained) speech models across linguistic categories. We evaluate probing accuracy at each layer of speech models whose weights are randomly initialized (i.e., no pretraining). }
    \label{fig:layerwise_untrained}
\end{figure*}

\begin{figure*}[h]
    \centering
    \includegraphics[width=1.0\linewidth]{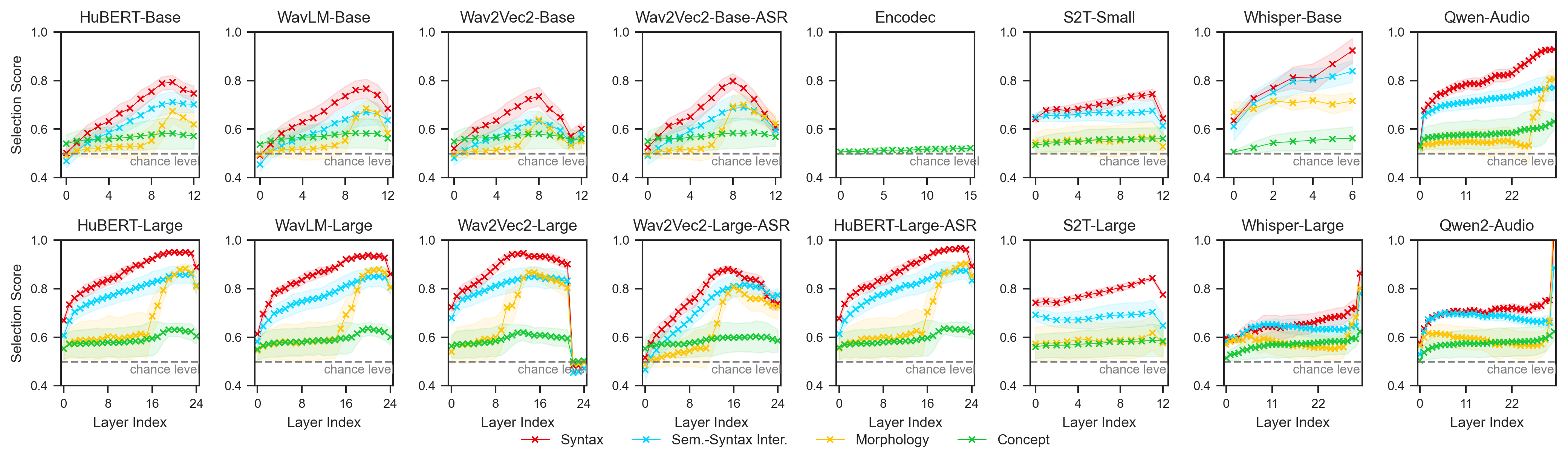}
    \caption{Layer-wise \textit{selection scores} reveal where each linguistic property is \emph{most informative}.}
    \label{fig:layerwise_selection}
\end{figure*}

\begin{figure*}[h]
    \centering
    \includegraphics[width=1.0\linewidth]{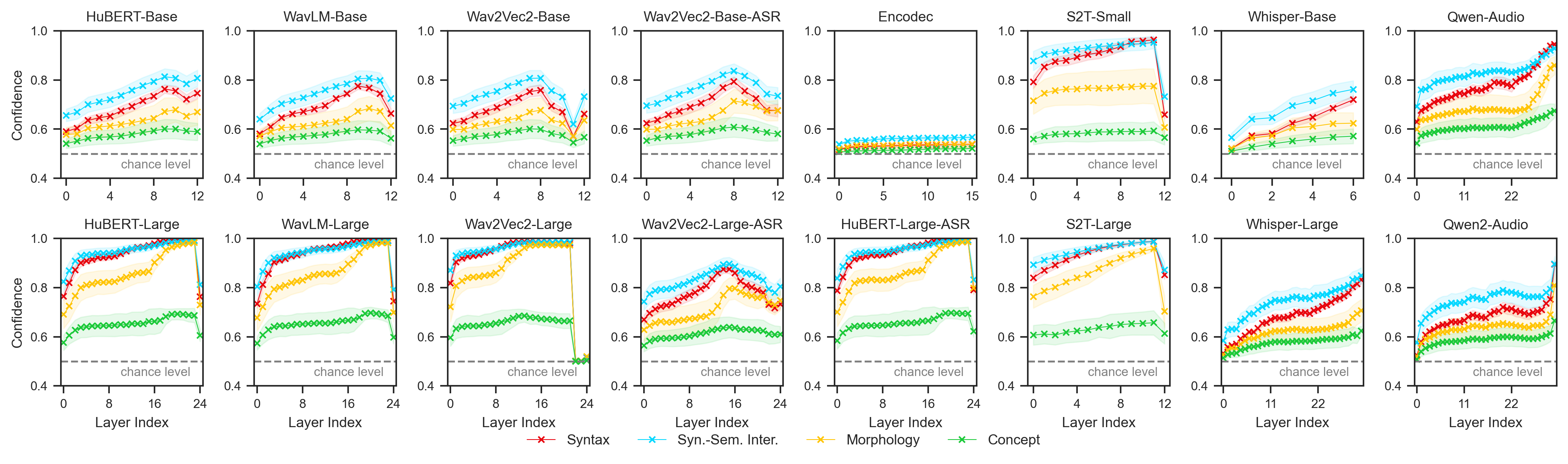}
    \caption{Layer-wise \textit{confidence scores} reveal how strongly models commit to correct predictions.}
    \label{fig:layerwise_confidence}
\end{figure*}
\paragraph{Untrained models' performance as baseline.} As shown in Figure~\ref{fig:layerwise_untrained}, despite the absence of learned weights, many models achieve above-chance performance, particularly for syntax and interface tasks. This suggests that input representations—derived from log-mel spectrograms or early convolutional layers—already encode structure-correlated acoustic cues. The mild upward trend in some shallow models (e.g., S2T, Whisper) further highlights that even without pretraining, architectural priors combined with spectrotemporal patterns can give rise to latent linguistic signals. In contrast, semantic and conceptual tasks remain near chance throughout, indicating their reliance on abstract, model-learned representations.
\paragraph{Selection score} As show in Figure~\ref{fig:layerwise_selection}, relative to the raw accuracies in Figure~\ref{fig:layerwise_f1}, selection scores sharpen the contrast between layers: (i) self-supervised models show a steady climb, peaking in the upper-middle layers before tapering, indicating that the diagnostic signal becomes both stronger and more concentrated with depth; (ii) ASR-fine-tuned encoders plateau earlier and exhibit a late-layer drop, consistent with a redistribution of capacity toward transcription; (iii) AudioLLM encoders (Whisper, Qwen) maintain rising curves to the final layer, suggesting continued accumulation of structurally relevant information; and (iv) the EnCodec baseline remains at chance, confirming that its representations lack discriminative linguistic content even after normalization.
\paragraph{Confidence score} As shown in Figure~\ref{fig:layerwise_confidence}, confidence increases consistently with depth across nearly all models, mirroring trends seen in accuracy and selection score. However, models vary in their \emph{calibration}: self-supervised models and AudioLLMs exhibit sharp late-layer confidence gains, whereas ASR-fine-tuned encoders often peak earlier and display more modest increases, especially for conceptual and morphological tasks. EnCodec shows low, flat confidence across the board, consistent with its poor accuracy. Overall, confidence dynamics highlight not just where linguistic features emerge, but how decisively they are encoded in the internal representation.
\end{document}